%% file: main.tex
\documentclass[runningheads]{llncs}
\usepackage{graphicx}

\usepackage[utf8]{inputenc}
\usepackage{pgfplots}
\usepgfplotslibrary{groupplots}

\usepackage{amsmath,amssymb} 
\usepackage{color}
\usepackage[width=122mm,left=12mm,paperwidth=146mm,height=193mm,top=12mm,paperheight=217mm]{geometry}

\usepackage{placeins}

\newcommand{\fig}[1]{Figure~\ref{fig:#1}}
\newcommand{\sect}[1]{Section~\ref{sect:#1}}
\newcommand{\tab}[1]{Table~\ref{tab:#1}}

\newcommand{\eq}[1]{(\ref{eq:#1})}

\begin{document}
\pagestyle{headings}
\mainmatter
\def\ECCV14SubNumber{2819}  

\title{Impostor Networks for \\ Fast Fine-Grained Recognition} 


\authorrunning{Vadim Lebedev, Artem Babenko, and Victor Lempitsky}

\author{Vadim Lebedev$^{1,2}$, Artem Babenko$^{2}$, and Victor Lempitsky$^1$}
\institute{$^1$Skolkovo Institute of Science and Technology (Skoltech), Moscow, Russia\\ $^2$Yandex, Moscow, Russia}

\maketitle

\begin{abstract} 
In this work we introduce impostor networks, an architecture that allows to perform fine-grained recognition with high accuracy and using a light-weight convolutional network, making it particularly suitable for fine-grained applications on low-power and non-GPU enabled platforms. Impostor networks compensate for the lightness of its ``backend'' network by combining it with a lightweight non-parametric classifier. The combination of a convolutional network and such non-parametric classifier is trained in an end-to-end fashion. Similarly to convolutional neural networks, impostor networks can fit large-scale training datasets very well, while also being able to generalize to new data points. At the same time, the bulk of computations within impostor networks happen through 
nearest neighbor search in high-dimensions. Such search can be performed efficiently on a variety of architectures including standard CPUs, where deep  convolutional networks are inefficient. In a series of experiments with three fine-grained datasets, we show that impostor networks are able to boost the classification accuracy of a moderate-sized convolutional network considerably at a very small computational cost. 
\end{abstract}

\input{intro}

\input{related}

\input{method}

\input{experiments}
\input{conclusion}

\section*{Acknowledgements}
This work is partially supported by the Ministry of Education and  Science of the Russian Federation (grant 14.756.31.0001)


\bibliographystyle{splncs}
\bibliography{refs}
\end{document}

%% file: intro.tex
\section{Introduction}

The ability to perform fine-grained recognition is one of the hallmarks of the recent progress in deep learning. The best of fine-grained classifiers \cite{Cai17,Zheng17,Kong17,He17}, however, use very deep convolutional networks (ConvNets) such as those based on the VGG-architecture \cite{Simonyan14}, which means that they are ill-suited for the deployment on mobile platforms and other platforms that lack GPUs, unless each image is processed remotely. At the same time, having a fine-grained classifier ``in your pocket'' and without the need for a remote server connection is what makes such classifiers particularly useful. 

A natural question is then, whether it is necessary to have a very big and deep ConvNet typically designed for large-scale visual recognition, in order to perform fine-grained classification? To address this question, in this work, we focus on building fine-grained classifiers that perform well and yet do not require a deep, computation- and power-hungry ConvNet to perform classification well.

Towards this goal, we suggest a new architecture that combines a compact ConvNet with a non-parametric classifier, and focus on squeezing a maximal performance from such combination. We argue that the combination is natural, as the non-parametric classifier is able to compensate for the inability of a compact ConvNet to achieve linear separation of similar visual classes.


The non-parametric classifier that is used in our system is a radial basis function classifier, which takes the high-dimensional output of an underlying ConvNet and then performs classification by combining proximity-based votes from a set of points in the embedding space. Our architecture is thus similar to the RBF-solver architecture recently proposed in \cite{Meyer17}, and also remininiscent of many works on metric learning (as several works evaluate k-NN classifiers on top of the learned metrics). In the above-mentioned approaches, the voting points are the mappings of the training examples by the learned embedding networks. The distinguishing property of our approach from both \cite{Meyer17} and metric learning approaches, is that in our case the voting points are not tied to the training samples. Instead the voting points are initialized to the images of the training examples under the ConvNet mapping, but drift away from such initialization as the learning progresses. We show that the extra flexibility resulting from the lack of ties between the training examples and the voting points results in a significant boost of the classification accuracy.

Our evaluation is performed on two popular fine-grained datasets (Caltech-UCSD Birds \cite{Wah11} and Stanford Cars \cite{Krause13}). To diversify the evaluation, we also perform experiments on the Landmarks-clean dataset \cite{Gordo16} of landmark images, where we again treat landmark recognition as a classification problem \cite{Li09}. In all cases, we observe that the classification accuracy of an underlying moderately-sized network (SqueezeNet \cite{Iandola16} in most of our experiments) is boosted considerably using our approach, while the additional computation cost is minimal. We also validate that the memory overhead of our approach over the baseline ConvNet can be decreased using standard compression schemes without affecting the classification accuracy strongly.

Finally, we evaluate the open-set ability of impostor networks, i.e.\ their ability to identify images that do not belong to training classes, and find that this ability also exceeds the standard classification networks.


The remainder of the work is organized as follows. In \sect{related}, we discuss the models and approaches related to impostor networks. In \sect{method}, we introduce impostor networks and their variants. We then present the experimental results in \sect{experiments}. We conclude with a short summary in \sect{summary}.

%% file: related.tex
\section{Related work} \label{sect:related}

\paragraph{Non-parametric image classifiers.} Non-parametric approach to classification (including image classification) has been popular for a long time. For example, kernel support vector machines \cite{KSVM} can be regarded as a non-parametric classifier (very much related to the RBF-classifier used in our method). Back at the outset of the ImageNet challenge, k-nearest neighbor classifier was suggested as a scalable approach for large-scale image classification \cite{Deng10}. It was, however, quickly outpaced by linear classifiers applied, firstly, on top of Fisher vectors \cite{Perronnin07} and later on top of the stack of deep learning feature hierarchies \cite{Krizhevsky12}. From that point, image classification was almost invariably tackled by progressively deeper and wider convolutional networks with linear classifiers at the top, in which the large number of underlying layers ensured the linear separability of classes at the last layer. Because of the sheer size, such architectures are almost always deployed on massively-parallel GPU architectures. Here, we focus on developing an image classifier that benefits from the deep learning advances and yet can run efficiently on non-massively parallel architectures.

\paragraph{Deep metric learning.} Our approach is naturally related a large and rapidly growing body of works on deep metric learning (DML) that train deep convolutional networks to produce embeddings that work well for nearest-neighbor retrieval \cite{Chopra05}. The method \cite{Salakhutdinov07} uses a neighborhood-component analysis loss \cite{Goldberger05} similar to the one used in our work. Some of the deep metric learning works use loss functions that implicitly or explicitly perform non-parametric estimation of class densities and maintain a list of representative class prototypes in the embedding space. This is e.g.\ true for magnet loss \cite{Rippel16} or proxy loss \cite{Movshovitz17}. At the same time, the number of such representatives is always much (orders of magnitude) smaller than the number of training examples, whereas the number of impostors in our approach equals the number of training images. The recently proposed radial basis solver approach \cite{Meyer17} is the closest to ours (and our approach was inspired by the experiments reported in \cite{Meyer17}). Their DML approach uses the same loss function \cite{Goldberger05}, and also evaluates the learned embeddings for classification with success.

Overall, while the combination of a deep convolutional network and a nearest neighbor search procedure makes DML methods similar to our method, the task that our system is trained for is different (classification of the classes seen at train time), which justifies the use of a different objective and the separation of the impostor set from the actual embeddings of the training examples (not performed in all DML methods including \cite{Meyer17}). Note, that while DML approaches can be used together with k-nearest neighbor classifiers, such classifiers have been reported to perform worse than standard softmax-loss based classifiers \cite{Rippel16,Sohn16}.   

\paragraph{Speeding-up ConvNets} has been a topic of very active research. The ideas include using low-bit quantizations of weights and activations \cite{Hubara16,Bagherinezhad16}, various tensor factorizations applied to convolutional and fully-connected layers \cite{Denton14,Lebedev14,Novikov15}, structured sparsification of convolutional kernels \cite{Lebedev16,Wen16}. All these methods are largely ``orthogonal'' to the approach introduced in our work and can be straightforwardly combined with it, as such methods can be used to speed-up ConvNets inside impostor networks. For this reason we do not compare with the above-mentioned speed-up methods, and use a popular lightweight ConvNet (SqueezeNet \cite{Iandola16}) through most of our experiments.




%% file: method.tex
\newcommand{\x}{{\mathbf x}}
\newcommand{\y}{{\mathbf y}}
\newcommand{\cY}{{\mathbf Y}}
\renewcommand{\c}{{\mathbf c}}
\newcommand{\loss}{\ell}

\section{Impostor Networks} 
\label{sect:method}

We now discuss our model. An \textit{impostor network} is an image classifier consisting of a convolutional network $f_\theta$ with learnable parameters $\theta$ that maps an input image $\x$ into a $d$-dimensional space $\cY$ as well as a dataset of reference points $\c_1,\dots,\c_M$ in $\cY$ with assigned class labels $l_1,\dots,l_M$ that define class kernel densities in $\cY$.

A trained impostor network classifies an image $\x$ by first mapping it to $\cY$:
\begin{equation} \label{eq:convnet}
\y = f_\theta(\x)
\end{equation}
and then computing a set of weights $w_1,\dots,w_M$ using the Gaussian kernel $g(\cdot,\cdot;\sigma)$ traditionally used in the RBF networks:
\begin{equation} \label{eq:kernel}
    w_j = g(\y, \c;\,\sigma) = \exp\left(-\frac{\|\y - \c_j\|^2}{2\sigma^2}\right)\,,
\end{equation}
where $\sigma$ is the standard deviation of the used kernel, which serves as a meta-parameter of the model. Within our approach, $\sigma$ is set by validation (although it can be included into the gradient-based learning formulation).
The classification process then predicts the probability of the image $\x$ belonging to a certain class $l$ using the radial basis function prediction rule:
\begin{equation} \label{eq:rbf}
    p(l(\x)=l) = \frac{1}{\sum_{j=1}^M  g(\y,\c_j;\sigma)} \sum_{j=1}^M  g(\y,\c_j;\sigma) \left[l_j=l\right]\,,
\end{equation}
where $[l_j=l]$ is an Iverson bracket.

The parameters of the embedding network and the reference set are obtained from training data given in the form of training examples $\x_1,\dots,\x_M$ with labels $l_1,\dots,l_M$. In our approach, we introduce the same number of reference points $\c_1,\dots,\c_M$, and associate each training example $\x_j$ with the reference point $\c_j$ (the particular ways of such association are discussed below). The label $l_j$ is retained by the reference point $\c_j$, and thus is used after the training to classify new examples according to the RBF classification rule \eq{rbf}. Since each $\c_i$ serves as a certain representative of the training example $\x_i$ at test time, we call $\c_i$ an \textit{impostor} and the resulting architecture an \textit{impostor network}.

\begin{figure}
\centering
\includegraphics[width=0.8\textwidth]{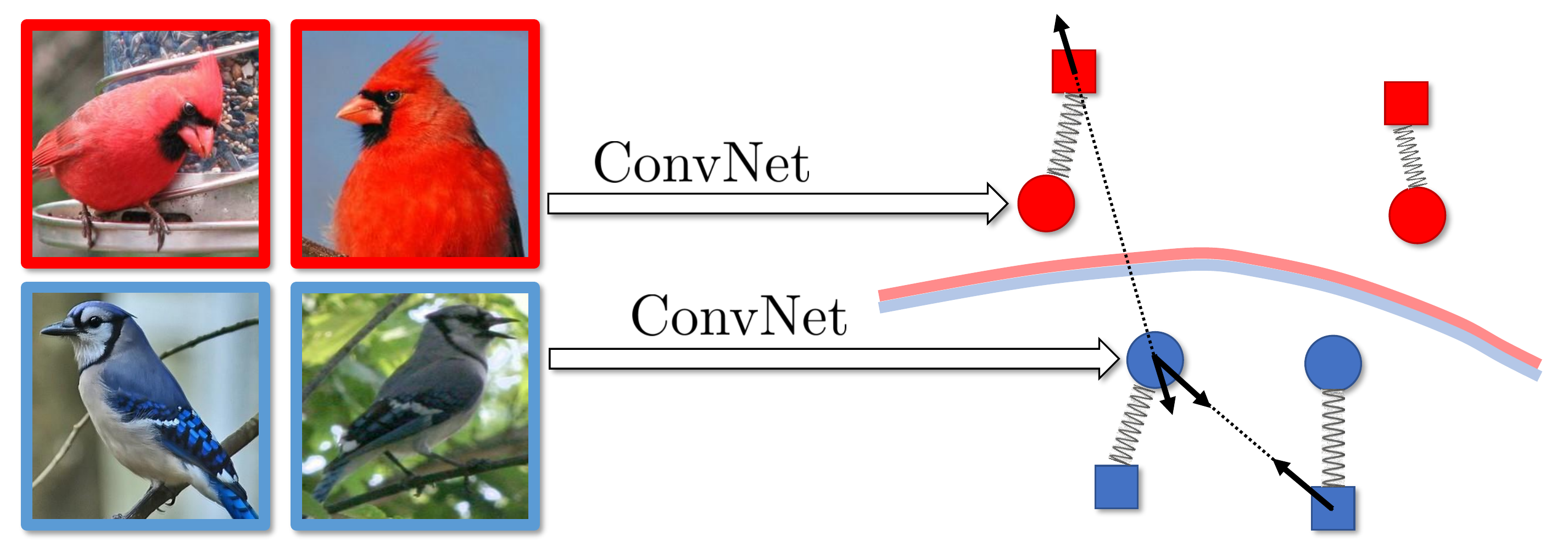}
\caption{The schematic view of the ''loose'' impostor networks training. The trainable parameters include the ConvNet weights (that determine the mapping from training images to their embeddings shown as circles) and the set of impostor vectors (shown as squares). Each impostor possess a class label, denoted by its color. Gradients of some loss terms are shown with arrows. They include the attraction term between an embedding of a training image and an impostor of the same class, as well as the repel term between an embedding of a training image and an impostor of another class. Additionally, loss terms penalize the deviation between the corresponding impostors and embeddings (shown with strings). At test time, the non-linear decision boundary of the classifier is determined by the position of impostors of various classes.}
\end{figure}

\subsection{Motivation}
The estimation of class probabilities using the rule \eq{rbf} can be performed efficiently even for a large number $M$ given that the bandwidth $\sigma$ is sufficiently small. Under this condition, approximate nearest neighbor search can be used to retrieve an (approximate) set of close neighbors of the embedding vector $\y$, for which the weights $g(\y,\c_j;\sigma)$ are not very close to zero. For a reasonably small $M$ (in practice upto several tens of thousands, as in our experiments), even an exhaustive computation of all weights in \eq{rbf} constitutes a small fraction of the computation time, whereas the majority of the inference time is spent on the ConvNet computation in \eq{convnet}.

Compared to a standard ConvNet classifier, which utilizes linear classification on top of the feature hierarchy, the convolutional network inside the proposed architecture is used in conjunction with a highly non-linear RBF-classifier \eq{rbf}. Consequently, given a reasonable set of impostors, much fewer convolutional layers may be needed for the embedding $f_\theta$ in order to fit the resulting non-linear decision boundary defined by \eq{rbf}. This explains why in the experiments below impostor networks are able to achieve higher classification accuracy compared to ConvNets with similar architecture.

\subsection{Training impostor networks}

There are several possible approaches how the parameters $\theta$ of the ConvNet $f_\theta$ and the impostor set $\c_1,\dots,\c_M$ can be learned from a set of training examples $\x_1,\dots,\x_M$. We now discuss these approaches.

\paragraph{Tied impostors.} In the approach based on tied impostors, the learning results in $c_i = f_\theta(\x_i)$, i.e.\ each impostor is tied to the embedding of the training example $x_i$. The learning process can then  be performed by minimizing the classification loss first introduced in \cite{Goldberger05}:
\begin{eqnarray}
    L(\theta,\c_1,\dots, \c_M) =&\frac{1}{M} \sum_{i=1}^M  
     -\log \frac{\sum_{j \ne i}  g\left(\strut f_\theta(\x_i,\c_j);\sigma\right)[l_j=l_i]}{\sum_{j \ne i}  g\left(\strut f_\theta(\x_i,\c_j);\sigma\right)} \,, \label{eq:nca}\\
     &\text{subject to}\qquad \c_i = f_\theta(\x_i) \label{eq:tied}
\end{eqnarray}
Each term in \eq{nca} approximates the probability estimate \eq{rbf} for the correct class label of a training example (where the corresponding impostor is bypassed to avoid severe overfitting). The learning formulation (\ref{eq:nca}-\ref{eq:tied}) is, in fact, very similar to the one proposed in \cite{Meyer17} (and can also be regarded as an approach for deep metric learning). In practice, enforcing the hard constraint \eq{tied} during the optimization is non-trivial. Indeed, to evaluate the terms in \eq{nca} for a minibatch requires one requires to compute embeddings $\c_j = f_\theta(\x_j)$ for all close neighbors of the minibatch members. To address this challenge, \cite{Meyer17} suggested to maintain the cached copies (effectively, the impostors) of the embeddings at all times, that gradually become obsolete, as the optimization proceeds. The cached copies are updated by resetting to the actual embeddings once in several epochs. Such resetting leads to some problems, as it effectively leads to abrupt and considerable change of the optimization objective, and may result in learning instabilities.

\paragraph{Fixed impostors.} A simpler alternative, which in our experiments proved surprisingly efficient, is to fix the impostors in the beginning of the optimization process and to never update them. Thus, given the initial ConvNet parameters $\theta_0$, every impostor $\c_i$ is initialized to $f_{\theta_0}(\x_i)$, and then the optimization of the objective \eq{nca} is performed over $\theta$, while $\c_i$ are excluded from the optimization, and the constraint \eq{tied} is disregarded. 

In our experiments, the initial state of the network parameters $\theta_0$ corresponds to the result of the training on the well-known ILSVRC classification task. Given a good initialization $\theta_0$, the initialized impostor set $\c_1,\dots \c_M$ are likely to create reasonable decision boundaries in the high-dimensional space $\cY$, which may be simpler to fit for the ConvNet $f_\theta$ than to achieve linear separability of the classes. This explains why such a simple scheme, when initialized to a pretrained $\theta_0$, can outperform the standard ConvNet (starting with the same pre-initialization) considerably. 

\paragraph{Loose impostors.} The third scheme that we consider, can be seen as a generalization of the tied impostor scheme, which avoids its pitfalls. Here the impostor set is once again made a part of the optimization, however the hard constraint \eq{tied} is replaced with the soft penalty that drives the deviation between $\c_i$ and $f_\theta(\x_i)$ down, leading to the following learning formulation:
\def\largestrut {\rule[-.3\baselineskip]{0pt}{1.5\baselineskip}}
\begin{equation}
    L(\theta,\c_1,.., \c_M) = \frac{1}{M} \sum_{i=1}^M \left( \largestrut \right.  \lambda\,\|f_\theta(x_i) - \c_i\|^2  
     -\log \frac{\sum\limits_{j \ne i}  g\left(\strut f_\theta(\x_i,\c_j);\sigma\right)[l_j=l_i]}{\sum\limits_{j \ne i}  g\left(\strut f_\theta(\x_i,\c_j);\sigma\right)}\left.\largestrut \right).\label{eq:looseloss}
\end{equation}
Here, the parameter $\lambda$ controls the relative weight of the attachment loss. We have found that the performance of the method is rather insensitive to $\lambda$ and set it to $1$ in our experiments. The loss \eq{looseloss} is differentiable w.r.t.\ all parameters of the impostor network, including both the ConvNet parameters and the impostor positions. One can therefore use standard stochastic gradient-based techniques to minimize it. 

During a single training epoch, every impostor $\c_i$ participates in the classification of training samples multiple times. Each time, the partial gradient of the loss \eq{looseloss} with respect to $\c_i$ is computed and the impostor position $\c_i$ is updated accordingly (i.e.\ pulled towards the embeddings of the training examples of the same class or pushed away from  the embeddings of the training examples of different classes). Once during every epoch, when the training example $\x_i$ is included in the mini-batch, the impostor $\c_i$ is also pulled towards the embedding $f_\theta(\x_i)$. 

In the loose impostor formulation, untying the impostors from the embeddings of the training examples as well as from their initial approximations greatly increases the capacity of the model \textbf{without} increasing its computational complexity at test time. This is because the coordinates of impostors effectively become learnable parameters. This may both have a beneficial effect of decreasing underfitting and the negative effect of increasing overfitting.
Also, compared to the tied impostors scheme, the gradients of the loss function in the loose scheme are computed without ignoring any terms and without cached approximations, making learning process more stable. Compared to the fixed impostor scheme, the process can potentially improve the decision boundary (if the initialization $\theta_0$ is highly sub-optimal).  In the experimental section, we compare all three impostor learning schemes.



\paragraph{Impostor compression.} One obvious downside of impostor networks compared to standard ConvNets is that the impostor set containing $M$ $d$-dimensional embeddings has to be distributed as a part of the network and maintained in the memory during operation. It is natural therefore to consider compression schemes for this set. One important realization is that in the fixed impostor scheme the embedding network $f_\theta$ can adapt to the impostors corrupted by lossy compression. Thus, we consider the variant of the fixed impostor scheme when the impostors are compressed by Product Quantization (PQ)\cite{Ge14}. In the experiments we demonstrate that PQ radically reduces the memory consumption of the impostor networks without noticeable drop in the classification accuracy.

%% file: experiments.tex
\section{Experiments}
\label{sect:experiments}

In this section we present the experimental evaluation of the proposed impostor networks.

\textbf{Datasets.} We perform the experiments on three fine-grained categorization datasets:

\begin{enumerate}
    \item Caltech-UCSD Birds dataset (CUB-200-2011)~\cite{Wah11}, containing 11,788 bird images (5994 train and 5794 test images) of 200 classes.
    \item Stanford Cars dataset~\cite{Krause13}, containing 16,185 car images (8144 train and 8041 test images) of 196 classes.
    \item Landmarks-clean dataset~\cite{Gordo16}, containing 35423 images of 671 different landmarks (30837 train and 4586 test images). While usually used with landmark protocols, the task of landmark recognition may also be treated as a classification problem \cite{Li09}.
    \item Fungi 2018 dataset (part of 2018 version of iNaturalist~\cite{Horn17} contest) contains 85578 training images and 4182 validation images of 1394 fungi species.
\end{enumerate}

For training the network we split all the datasets into train/validation/test subsets and use the validation subset to tune the metaparameters (including $\sigma$). Our main performance measures are the standard classification accuracy and the inference time on both CPU and GPU.




\textbf{Experimental details.} 
During both training and applying the networks, we resize the input images to $256\times256$. When training, random cropping and mirroring were also applied. For the Cars and Birds datasets, we follow \cite{Meyer17} and use the provided bounded boxes to crop the object of interest in the image. Learning rates for all the schemes and the $\sigma$ parameter of the RBF-classifier in the impostor networks were tuned on the validation subsets. For training, we use the Adam optimizer\cite{Kingma14} with the weight decay parameter equals $5\times10^{-4}$. We train all the models for $60$ epochs. All  experiments were performed in PyTorch~\cite{paszke2017} framework, and we will release our code and models upon publication.


The protocol for training the impostor networks includes the following steps:
\begin{enumerate}
    \item We initialize the ConvNet with weights, obtained from the pretraining on ILSVRC. If the desired embedding dimensionality $d_{em}$ does not equal to the dimensionality of the last fully-connected layer $d_{fc}$, we change the size of the last fully-connected layer to be $d_{fc}\times d_{em}$, and  initialize it by a random matrix. 
    \item We pass all the train images through the ConvNet and use the outputs of the last fully-connected layer as initial impostor vectors $c_1,\dots,c_M$.
    \item We divide the impostors and the values of the matrix in the last layer of the network by the average impostor $L2$-norm $\frac{1}{M}\sum_{i=1}^M||c_i||$. This trick is important in practice, as the typical scale of distances between impostors varies greatly with $d_{em}$ and the scale of the last layer parameters. Due to this variability, the $\sigma$ parameter has a very wide range of possible values, which makes it hard to tune. The division by the average norm allows the scale of distances to stay the same across different $d_{em}$ and initializations.
    \item The standard backpropagation is applied to minimize the corresponding impostor network loss: \eq{tied} or \eq{looseloss}.
\end{enumerate}

Due to a small number of images in the evaluation datasets, the RBF classification rule is computed with exhaustive search, i.e. distances to all impostors are calculated.

\textbf{Compared approaches.} The state-of-the-art methods for fine-grained recognition\cite{Simon17,Zheng17} focus solely on the classification accuracy and it could take up to dozens of seconds to use them on mobile platforms. In this paper we target the demanding operating point of ''lightweight'' approaches. In particular, we consider the methods and network architectures that could be employed on non-GPU platforms and used in realtime, i.e. the inference should be faster than 10 FPS on CPU devices. Given these limitations, we compare the following schemes: 
\begin{enumerate}
    \item \textbf{ConvNet:} for this baseline we take the ILSVRC-pretrained network and fine-tune it on the particular dataset using the standard cross-entropy loss. The inference in this scheme is very efficient as it requires the only forward pass.
    \item \textbf{ConvNet-extra:} to verify that simply adding more parameters into the ConvNet would not result in the accuracy boost, we compute this additional baseline, which mimics the ConvNet variant, except that an extra fully-connected layer with $d_{em}$ neurons is inserted before the last layer
    \item \textbf{ImpostorNet-tied:} the impostor network with impostors tied to the embeddings of the training examples. We follow \cite{Meyer17} and recompute the impostors every tenth epoch. Resetting impostors more frequently (e.g.\ after every epoch) was also tried but resulted in a worse performance.
    \item \textbf{ImpostorNet-fixed:} the impostor network with impostors fixed to the initial positions determined by the ILSVRC-pretrained network.
    \item \textbf{ImpostorNet-loose:} the impostor network with loose impostors that are part of the optimization process. Here, we stick t the value $\lambda=1$ (in our initial experiments we observed that changing this parameter does not improve the final result).
\end{enumerate}
All the networks in the compared schemes are initialized by the weights from the ConvNet, pretrained on the ILSVRC dataset\cite{Russakovsky15}. We always initialize the train images' impostors by the corresponding image embeddings, obtained with the initialization weights. For one of the three datasets, we have evaluated the following network architectures: SqueezeNet\cite{Iandola16} (PyTorch SqueezeNet version $1.1$ was used), AlexNet\cite{Krizhevsky12}, ResNet-18 and ResNet-50\cite{He15}. As SqueezeNet stood out in terms of accuracy/efficiency trade-off in this comparison, we used this architecture for the remaining two datasets. Unless noted otherwise, the embedding dimensionality for ImpostorNets is set to $512$.




\begin{table}[]
    \centering
    \caption{Classification accuracy on the Birds, Cars and Landmarks datasets for three versions of impostor networks and the ConvNet trained with cross-entropy loss (including the variant with additional learnable parameters). Impostor networks provide a substantial performance improvement on the Birds and Cars datasets in the case of SqueezeNet architecture.}
    \begin{tabular}{|c|c|c|c|c|c|c|}
        \hline
          & \multicolumn{4}{|c|}{Birds}& Cars & Landmarks \\
        \hline
          & squeezenet & alexnet & resnet18  & resnet50 & squeezenet & squeezenet\\
        \hline
        ConvNet &  70.72 &  65.58 & 79.61 & 82.35 & 78.65 & 92.48 \\
       \hline
        ConvNet-extra & 69.30 & 65.28 & 80.08 & 82.58 & 76.67 & 92.67 \\
        \hline
        tied & 70.04 & 62.10 & 75.68 &  80.54 & 80.84 & 93.65\\
        \hline
        fixed & 75.47 & 66.97 & 80.05 & 82.68 & 79.87 & 93.02\\
        \hline
        loose & 76.01 & 67.30 & 80.58 & 82.14 & 85.05 & 92.83\\
        \hline
    \end{tabular}
    \label{tab:my_label}
\end{table}

\textbf{Classification accuracy.} We demonstrate the classification accuracy for all the compared methods in \tab{my_label}. We highlight several key observations from it:
\begin{enumerate}
    \item For the Birds dataset, where we compare different architectures, the advantage of impostor networks is the most substantial for the most efficient Squeezenet architecture. For the more powerful architectures the advantage of our scheme is much smaller. We believe that the reason of such behaviour is that the architectures with large number of parameters are able to make different classes linearly separable and the non-linear decision boundary is less useful.
    \item The version with loose impostors outperforms the baseline and versions with tied and fixed impostors in several cases and never performs much worse. We attribute this fact to the extra learning capacity of this scheme, explain it by the fact that in the ''loose'' version the impostors are trained jointly with the networks, hence many more model parameters adapt to the particular dataset.
    \item The advantage of impostor nets cannot be explained simply by having extra parameters in the decision function. In fact, adding more parameters into the ConvNet does not necessarily improves the performance (ConvNet-extra vs.\ ConvNet).
    \item The advantage of impostor nets on the Landmarks-clean dataset is small. This maybe due to already saturated performance on this dataset.
\end{enumerate}

We further investigate ImpostorNets with additional experiments.

\textbf{RBF-SVM baseline.} To prove the necessity of joint learning of ConvNet weights and impostor positions in embedding space, we compare ImpostorNets with RBF-SVM classifier trained on fixed feature vectors extracted from the last hidden layer of ConvNet. This experiment was performed for Squeezenet architecture on Birds dataset and yielded accuracy of 69.2$\%$, lower then original ConvNet and every ImpostorNet version.

\textbf{Fungi 2018.} We have performed preliminary experiments on this Fungi 2018 to confirm the applicability of impostor networks for larger datasets. Impostor networks in the ''loose'' version with SqueezeNet architecture achieved accuracy of 26.6$\%$, compared with 25.7$\%$ of original ConvNet.

\begin{figure}
\centering
\label{fig:embed_dim_plot}
\newlength\figureheight
\newlength\figurewidth
\setlength\figureheight{3cm}
\setlength\figurewidth{3cm}
\begin{tabular}{cc}
\input{plots/dim.tex}&
\input{plots/compression.tex}
\end{tabular}
\caption{Left -- the classification accuracy of ''fixed'' impostor networks on the Birds dataset as a function of the impostors dimensionality $d_\text{em}$. Larger $d_\text{em}$ results in higher performance but also increases the memory consumption and the computational complexity of RBF classification (although the computational cost still remains very small compared to the network inference time). Right -- same accuracy as a function of the compressed impostor representation size (which is given in bytes). The impostors could be compressed to as little as $16$ or $32$ bytes with negligible accuracy drop compared to uncompressed impostors.}
\end{figure}
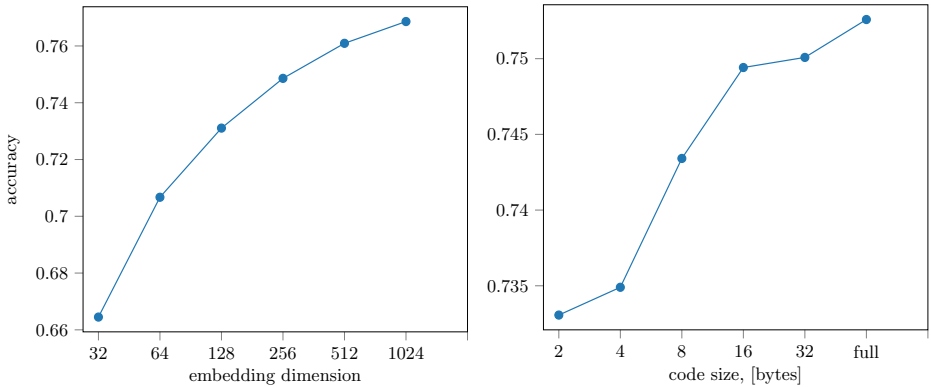

\textbf{Dependence on the embedding dimensionality $\mathbf{d_{em}}$.} The dimensionality of the embeddings and the impostors $d_{em}$ has a large influence of the impostor networks performance. \fig{embed_dim_plot} demonstrates the classification accuracy on the Birds dataset as a function of $d_{em}$ for the ''fixed'' version of the impostor network. As expected, larger dimensionalities result in more powerful models (due to larger number of parameters). On the other hand, the large $d_{em}$ values increase the memory consumption and RBF classification complexity. In most of our experiments we set $d_{em}$ to $512$.


\textbf{Initialization variants.} In the ''fixed'' version of our scheme the impostors are not updated during training and are fully defined by the initialization weights. In this experiment we compare several different ways to initialize the ''fixed'' impostors with the SqueezeNet architecture. The initialization with the weights, obtained with pretraining on the ILSVRC dataset, results in $76\%$ classification on the Birds dataset. As expected, the random impostors initialization results in poor classification performance of $16.1\%$. Another reasonable way is to initialize the impostors with the weights obtained after finetuning on the Birds dataset with cross-entropy loss. We observed that this initialization results in slightly higher final performance of $76.9\%$. However, this increase in the accuracy comes at a cost of the additional ConvNet training.



\textbf{Impostors compression.} Here we investigate the performance of the ''fixed'' version of our approach when impostors are represented by the compact PQ codes to reduce memory consumption. In particular, we use the optimized version of PQ, proposed in \cite{Ge14}, for compression. \fig{embed_dim_plot}--right demonstrates the classification accuracy on the Birds dataset as a function of the code size (presented in bytes). In this experiment we initialize the impostors, compress them with OPQ, and then train the impostor network using the compressed vectors as the impostor set $c_1,\dots,c_M$. The graph shows that the original $512$-dimensional impostors could be compressed to $16$ or $32$ bytes with only negligible accuracy drop. Note, that the usage of PQ codes for impostors does not prevent the efficient RBF classification, as the distances to the compressed impostors could be computed efficiently both on CPU and GPU~\cite{Johnson17}.
Overall, the total memory consumption of the impostor networks is dominated by the memory required to store CNN weights if the impostors are PQ-compressed. E.g. for the Birds dataset, the size of the SqueezeNet model is about 4.8MB, while the impostors, compressed to $16$ bytes, require only 92KB of additional memory.



 






\subsection{Timings}

As in this paper we focus on the efficient architectures for fine-grained recognition, we measure the inference timings for the compared approaches both on CPU and GPU. The GPU timings are recorded on single Tesla K40m with CUDA 8.0, and for CPU timings we use Intel Xeon CPU E5-2650 v2 2.60GHz.

\begin{figure}
\centering
\label{fig:timings_plot}
\includegraphics[width=\linewidth]{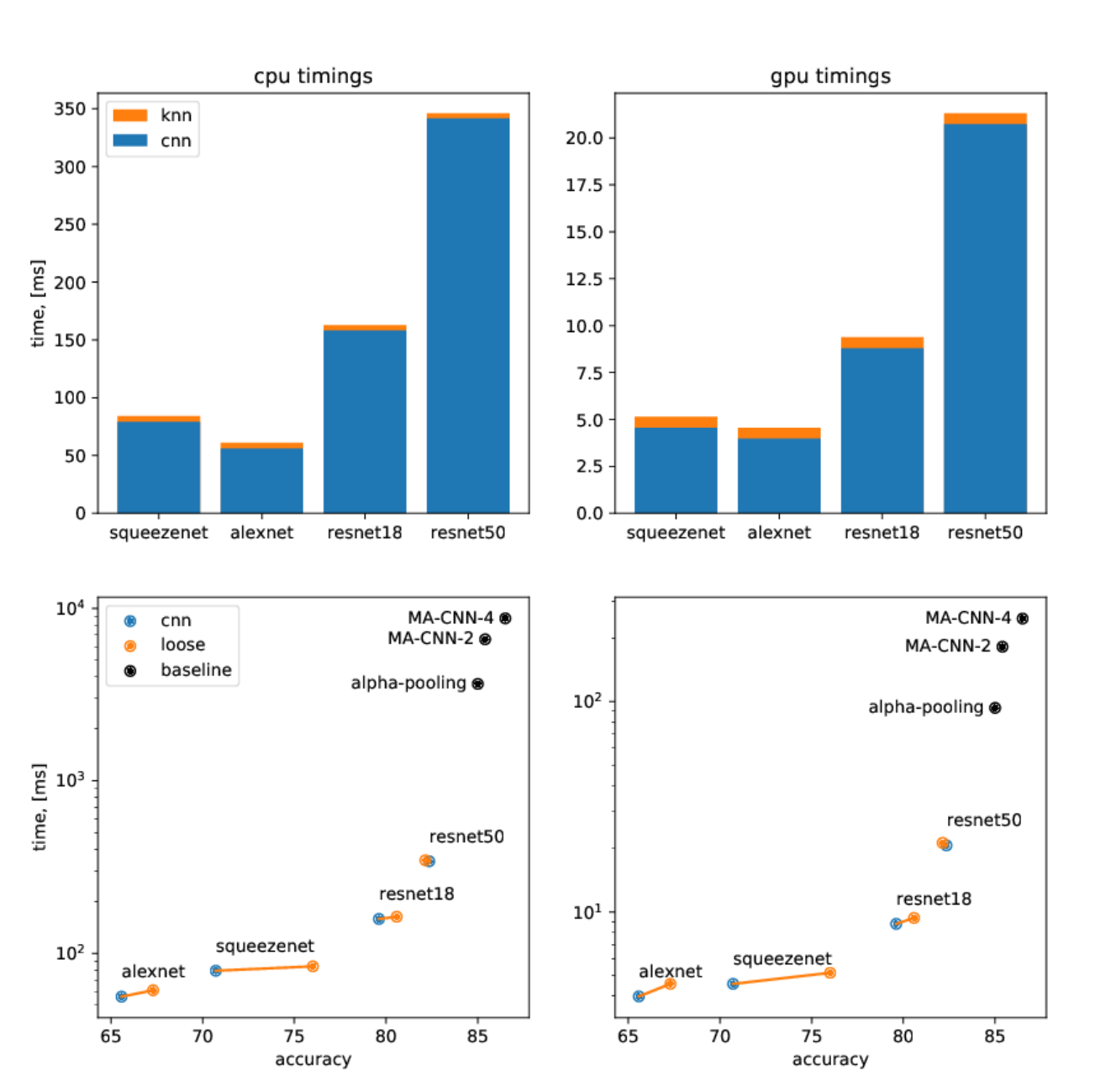}
\caption{Top panels: the CPU and GPU timings for the impostor networks with different ConvNet architectures on the Birds dataset. Blue bars correspond to the computational times of ConvNet forward pass, while orange bars correspond to the RBF classification rule computation. The RBF contribution into the total runtime is negligible for all architectures. Bottom panels: comparison of the ''loose'' impostor networks with the state-of-the-art approaches in terms of runtime-accuracy trade-off on the Birds dataset. The proposed impostor networks are orders of magnitude faster the the state-of-the-art methods with decent decrease in the classification accuracy.}
\end{figure}

Note, that the state-of-the-art approaches for fine-grained classification rarely take inference timings into consideration and do not report them. To position our approach among the previous works, we also compute the timings for two recent approaches, which have their models publicly available~\cite{Simon17,Zheng17}. \fig{timings_plot} shows the timings, achieved by our system as well as the timings of the state-of-the-art approaches. All the timing are computed on the Birds dataset.

The top part of \fig{timings_plot} demonstrates the timings of impostors networks with different architectures. We report the timings of ConvNet forward pass and distances computation separately to demonstrate that the contribution of the latter is quite small for all architectures, even with exhaustive nearest neighbor search. Of course, for larger training sets exhaustive search could be inefficient and the approximate methods should be used.

The bottom parts of \fig{timings_plot} compares the ''loose'' version of our system with the existing approaches in terms of time-accuracy trade-off on the Birds dataset. The black points correspond to the performance of two recent methods, $\alpha-$pooling\cite{Simon17} and multiple attention with different parameters (MA-ConvNet-2 and MA-ConvNet-4)\cite{Zheng17}. The blue points correspond to the performance of ConvNet trained with the standard cross-entropy loss. Finally, the orange points denote the performance of our system (loose impostors).

In general, the proposed impostor networks improve the classification accuracy with negligible increase in the computational cost. The improvement is most noticeable for the ''lightweight'' SqueezeNet architecture, what makes our system a good choice for mobile platforms. Note, the state-of-the-art methods \cite{Simon17,Zheng17} achieve higher classification accuracy, but their runtimes are orders of magnitude slower on both CPU and GPU, which restricts its usage in practice.




\subsection{Open set recognition.} In this experiment we demonstrate that the proposed impostor networks could be successfully used in the ''open set learning'' scenario, when the images of the classes unseen during training could be submitted to a network at test time. The ability to detect the inputs of unknown classes is an important practical property for many application scenarios of fine-grained recognition. Here, we investigate the ability to detect unknown classes based on the confidence (entropy) of network predictions.

\begin{figure}
\centering
\label{fig:entropy_hist}
\includegraphics[width=\linewidth]{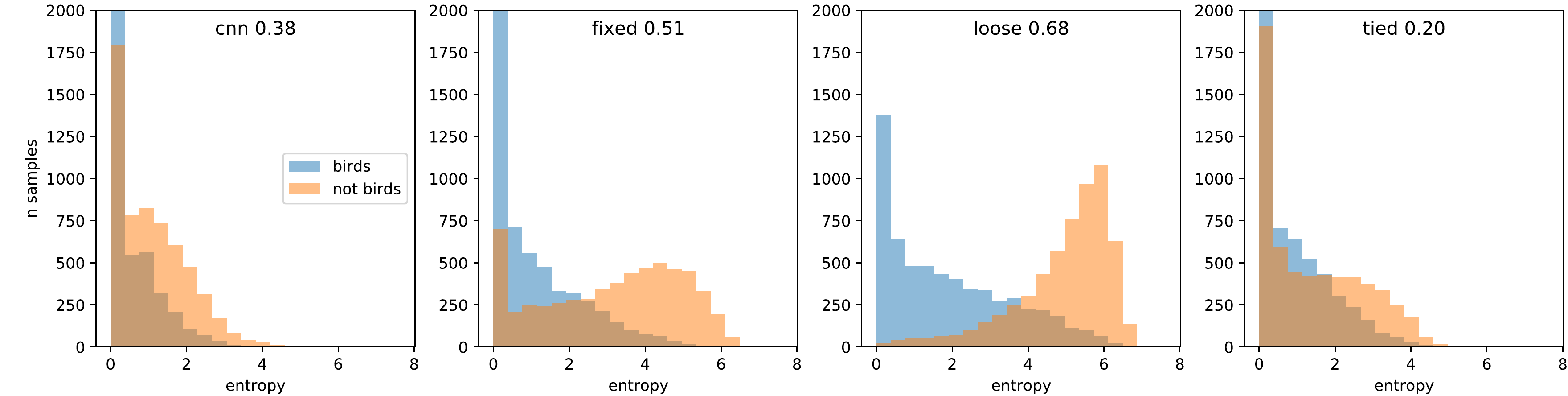}
\caption{Histograms of entropy values of the probability distributions obtained with the cross-entropy ConvNet and the impostor networks with the Squeezenet architecture. All the networks were trained on the Birds dataset and applied to the sets of bird and non-bird images. The cross-entropy ConvNet is almost equally confident on the images of both seen and unseen classes. The ''fixed'' and ''loose'' impostor networks are able to detect their uncertainty based on much higher entropy values on the non-bird images. The number on each plot corresponds to the Kolmogorov-Smirnov distance between the distributions of the entropy values.}
\end{figure}

It is well-known that the ConvNets trained with the standard cross-entropy loss tend to be overconfident even when they are wrong~\cite{Guo17}. Interestingly, the proposed impostor networks are able to express uncertainty due to the usage of RBF classification at the final step. To support this claim, we perform the following experiment. We train the cross-entropy ConvNet and the three variants of the impostor networks with the same SqueezeNet architecture on the Birds dataset. We then ran the models on two test sets: Test1 set containing birds images and Test2 set containing the non-bird images from the ILSVRC dataset. For each of these sets we pass the images through the networks and compute the entropy of the class labels probability distribution. The histograms of the entropy values are presented on \fig{entropy_hist}. For the standard ConvNet two distributions are very close, which means that the degree of the network confidence is the same for seen and unseen classes. The fixed and loose impostor networks clearly tend to be less confident in their predictions for the images of unseen classes.

\begin{figure}
\centering
\includegraphics[width=\linewidth]{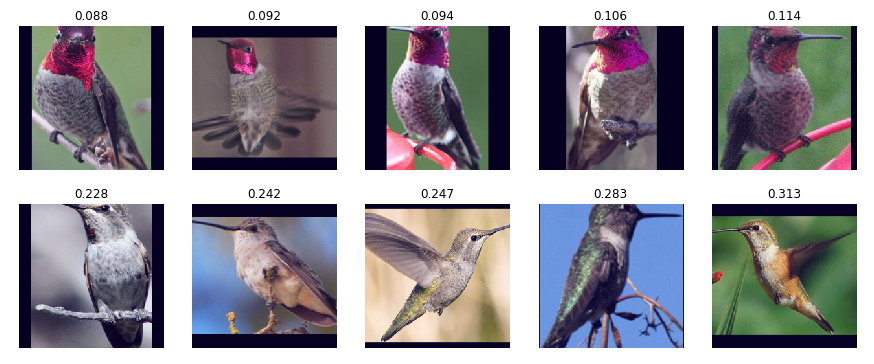}
\caption{Examples from Anna's hummingbird class of the Birds dataset with the smallest (top) and the largest (bottom) distances between their embeddings and the corresponding impostors for trained ''loose'' impostor network. The largest distances correspond to samples that are generally hard to classify. Anna's hummingbirds exhibit pronounced sexual dimorphism: males have characteristic bright magenta crown, while females are much more bleak. Indeed, all the specimen on the top row are males and on the bottom row --- females.}
\label{fig:tenbirds}
\end{figure}

\subsection{Intuition behind the ''loose'' impostors}

Finally, we perform an experiment that allows to obtain an intuition about mutual positions of the image embeddings and the corresponding ''loose'' impostors in the embedding space. In this experiment we take the training images of Anna's hummingbird class of the Birds dataset and compute the distances between their embeddings and the corresponding impostors. The bottom row of \fig{tenbirds} demonstrates top-5 images with the largest embedding-impostor distances. These images are hard to classify as they exclusively contain female hummingbirds, which lack the distinctive coloring of male Anna's hummingbirds. The top row of \fig{tenbirds} visualizes top-5 images with the smallest embedding-impostor distances.  These images correspond to easy samples of brightly coloured male hummingbirds, with embeddings which are far away from the decision boundary in the embedding space.

%% file: plots/dim.tex
\begin{tikzpicture}[scale=0.75]

\definecolor{color0}{rgb}{0.12156862745098,0.466666666666667,0.705882352941177}

\begin{axis}[
xlabel={embedding dimension},
ylabel={accuracy},
xmin=-0.25, xmax=6,
ymin=0.659292625959293, ymax=0.773807140473807,
xtick={0,1,2,3,4,5,6},
xticklabels={32,64,128,256,512,1024,},
tick align=outside,
tick pos=left,
x grid style={lightgray!92.026143790849673!black},
y grid style={lightgray!92.026143790849673!black}
]
\addplot [semithick, color0, mark=*, forget plot]
table {%
0 0.664497831164498
1 0.706706706706707
2 0.731064397731064
3 0.748581915248582
4 0.760927594260928
5 0.768601935268602
};
\end{axis}

\end{tikzpicture}

%% file: plots/compression.tex
\begin{tikzpicture}[scale=0.75]

\definecolor{color0}{rgb}{0.12156862745098,0.466666666666667,0.705882352941177}

\begin{axis}[
xlabel={code size, [bytes]},
xmin=-0.25, xmax=6,
ymin=0.73209042375709, ymax=0.753561895228562,
xtick={0,1,2,3,4,5,6},
xticklabels={2,4,8,16,32,full,},
yticklabel style={/pgf/number format/precision=3},
tick align=outside,
tick pos=left,
x grid style={lightgray!92.026143790849673!black},
y grid style={lightgray!92.026143790849673!black}
]
\addplot [semithick, color0, mark=*, forget plot]
table {%
0 0.733066399733066
1 0.734901568234902
2 0.74341007674341
3 0.749416082749416
4 0.750083416750083
5 0.752585919252586
};
\end{axis}

\end{tikzpicture}

%% file: conclusion.tex
\section{Summary}
\label{sect:summary}

In this paper we propose a new framework of impostor networks for efficient fine-grained classification. The impostor networks consist of the deep convolutional network with a non-parametric classifier on top. The ConvNet and the RBF parts are learned jointly and we investigate three possible ways to perform such joint learning. The ConvNet part could benefit the ConvNets of any architecture but the most benefits are achieved for the efficient SqueezeNet architecture what makes the impostor networks a good choice for resource-constrained settings, e.g. mobile platforms. 